# Automatic Skin Lesion Segmentation Using GrabCut in HSV Colour Space

Fakrul Islam Tushar

Erasmus+ Joint Master Program in
Medical Imaging and Applications
University of Burgundy (France),
University of Cassino (Italy) and
University of Girona (Spain)

f.i.tushar.eee@gmail.com

*Abstract*—Skin lesion segmentation is one of the first steps towards automatic Computer-Aided Diagnosis of skin cancer. Vast variety in the appearance of the skin lesion makes this task very challenging. The contribution of this paper is to apply a power foreground extraction technique called GrabCut for automatic skin lesion segmentation with minimal human interaction in HSV color space. Preprocessing was performed for removing the outer black border. Jaccard Index was measured to evaluate the performance of the segmentation method. On average, 0.71 Jaccard Index was achieved on 1000 images from ISIC challenge 2017 Training Dataset.

*Keywords—Skin lesion, segmentation, GrabCut, HSV, color space, Melanoma.*

## I. INTRODUCTION

Skin Cancer is one of the most rapidly increasing cancer all over the world with one in every three cancer diagnosed is a skin cancer according to the World Health Organization [1]. Malignant Melanoma a type of skin cancer estimated to have 76,389 new cases and over 100,00 deaths in the United States in 2016 [2]. So, Early diagnosis is very critical, as study showed survival rate for Melanoma increased over 90% if detected in the early stage [1]. Since skin cancer occurs at the surface of skin, visual inspection by a dermatologist using Dermoscopy is the common way for diagnosis.

Inspection of the dermoscopic images for dermatologist usually a complex and time-consuming task. To assist the dermatologist and improve the accuracy of the diagnosis computer-aided diagnosis systems have been developed. Skin lesion segmentation is very important part of the CAD systems for diagnosis. However, automatic skin lesion segmentation of skin lesion is very challenging due to large variety of appearance in color, texture, and size for different patients. In addition to these hair, veins medical gauzes and light reflections makes it more difficult task. Fig.1 Shown some example of the skin lesions.

In early years many literatures tried to solve this segmentation problem proposed methods based on mainly thresholding, Active contour, clustering and supervised learning. Authors in [3] used image enhancing using color and brightness saliency maps with Otsu thresholding. Garnavi et al. [4] used 25 different color channels for segmentation using thresholding. Region splitting and merging algorithm used in [5]. Authors in [6, 7] used the active contour methods for the segmentation. Melli et al. [8] used mean shift clustering. However, application of these methods was not very effective due to large difference in appearance of the skin lesion and requirement for proper preprocessing of the dermoscopic images.

Since last few years Convolutional Neural Networks (CNNs) become an obvious choice to the computer vision society [ 9,10]. Authors in [11,12,13,14,15] applied CNN for segmentation task and achieved high evaluations.

In this paper we demonstrate a simple yet powerful image segmentation technique GrabCut to segmentation the skin lesion using HSV color spaces impact on segmentation performance.

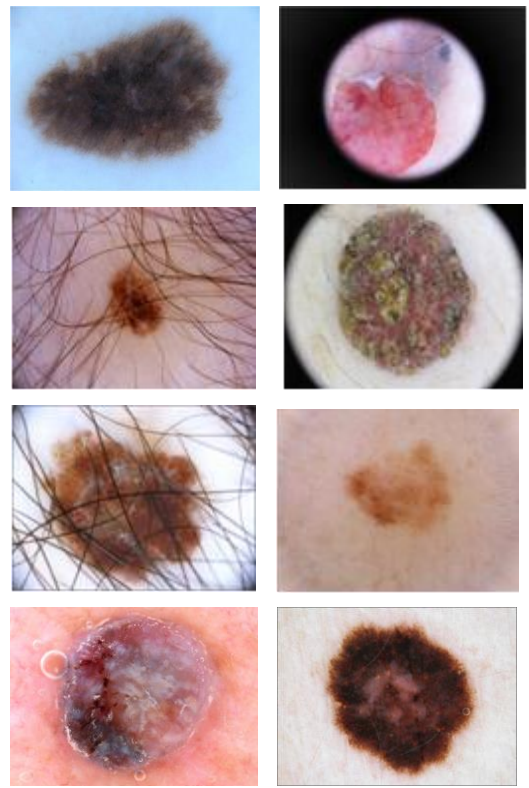

Fig.1. Sample Images from ISIC 2017 dataset.

## II. DATASET

The dataset used for this paper were download from the International Skin Imaging Collaboration (ISIC) 2017 Challenge. ISIC 2017 dataset contains training set of 2000 images with corresponding ground truth images. We have used 1000 images from the ISIC 2017 dataset in this paper.

## III. METHOD

GrabCut is one of the most powerful background and foreground extraction techniques that uses minimum graph cut for segmentation. Authors in [tushar-1] presented the algorithm. Using the edge and regional information of the given image an energy function was created and this function is being optimized depending on the binary label {0,1} provided by the minimum cut on the image graph.

In this paper an automatic modified version of GrabCut algorithm was used for skin lesion segmentation with minimal human interaction. Fig.2 represents the workflow of the pipeline used for segmentation.

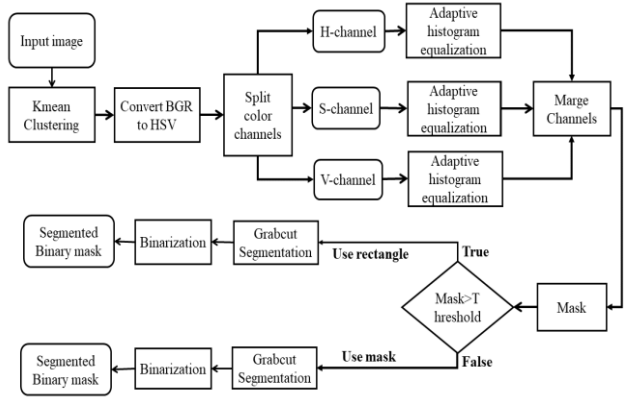

Fig.2. Overall pipeline of the segmentation.

### A. Preprocessing

Preprocessing was performed to remove the outer dark border from the microscopic images. Figure.3 shown the preprocessing steps. Mask was created for each image depending on dark pixel values. Afterward the mask was used to fill the dark border pixel with the neighbor pixels using inpainting. Pre-processed image was the input to the segmentation pipeline.

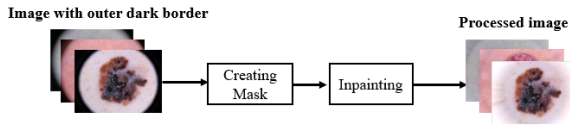

Fig.3. Dark outer boarder removal pipeline.

### B. Color Quantization

Color quantization was performed applying k-means clustering. Then the BGR image was converted to HSV color space. All the channels were split, and Adaptive histogram equalization was performed on each channel separately. Afterward all the color channels were marge.

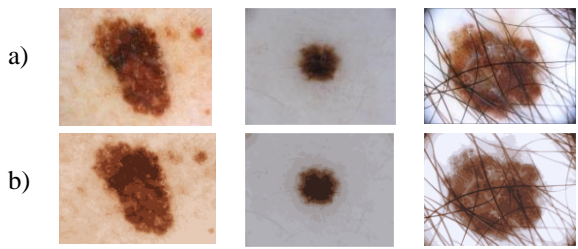

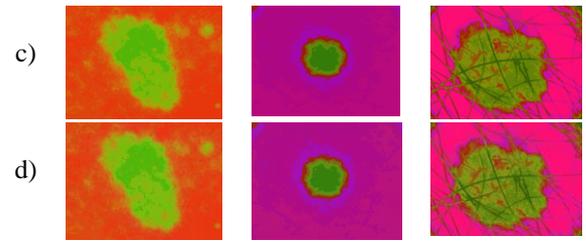

Figure.4 a) Original image. b) k-means clustering c) RGB to HSV and d) adaptive histogram equalization.

### C. Proposed Automated GrabCut

To perform GrabCut two approaches were used. First was using mask and second was using rectangle. In HSV color space most of the skin lesion appears to be green. Green color was extracted from the image and applying thresholding mask was obtained. Where ever the mask has value 1, image graph considered it as foreground and wherever was 0, considered as background. Then grabcut segmentation was performed.

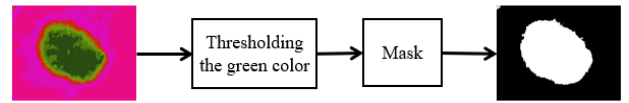

Fig.5. Mask creating workflow diagram.

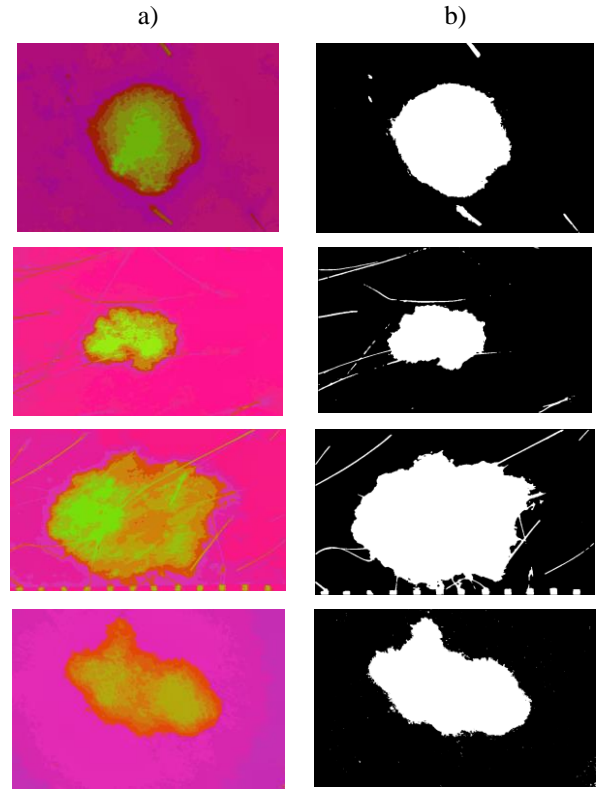

Fig.6. a) Enhance HSV image b) generated mask.

But in some cases, masking failed where very little intensity different from the skin and lesion.

To overcome this limitation, one thresholding approach was introduced. Threshold was calculated based on the mask dimension and intensity value. If the summation of intensity value exceeds and rectangle was generated. Equation (1) and Equation (2) shown the dimension of the rectangle.

Height of rectangle= height of image-(0.03×height of image)… (1)

Weight of rectangle= weight of image-(0.1×weight of image)…(2)

things outside the rectangle considered as background and GrabCut segmentation was performed.

## RESULTS

To evaluate the performance of the proposed segmentation method we used the evaluation matrix called Jaccard Index (JC). Jaccard Index (also known as Jaccard coefficient index) gives the similarity and diversity of sample sets.

Jaccard Index=TP/(TP+FP+FN) ……(3)

Here TP= lesion pixel segmented as lesion pixel, FP= non-lesion pixel segmented as lesion pixel, and FN= lesion pixel segmented as non-lesion pixel. We used ISIC challenge 2017 data applied proposed segmentation to 1000 images and achieved an average JC of 0.71. Some of the Segmented lesions by the proposed pipeline are given in Table. 1

TABLE I. Segmented skin lesion.

| Original Image | Ground Truth | Segmented Image | JC |
|---|---|---|---|
| 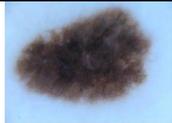 | 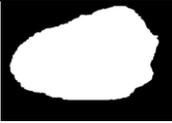 | 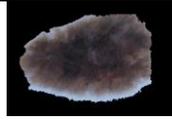 | 0.86 |
| 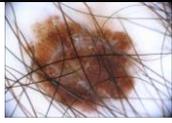 | 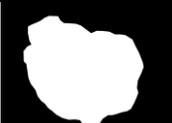 | 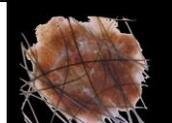 | 0.83 |
| 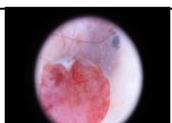 | 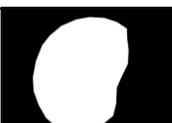 | 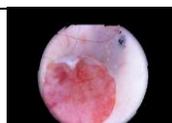 | 0.73 |
| 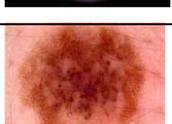 | 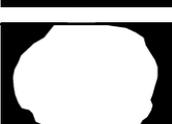 | 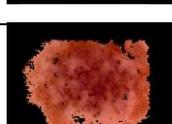 | 0.75 |
| 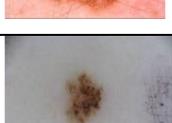 | 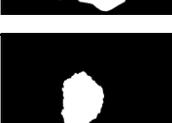 | 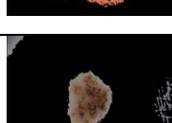 | 0.67 |
| 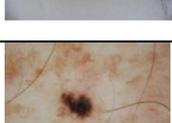 | 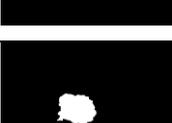 | 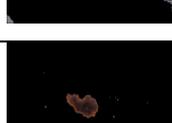 | 0.17 |

## DISCUSSION AND FUTURE WORK

Proposed segmentation technique is robust enough to segment the image without hair removal as shown in Table.1. Proposed pipeline performs comparable if the contrast between the skin lesion and non-lesion pixels are very less. The performance can be improved by applying more precise mask creating strategy, Applying different color spaces and addition the foreground probability estimation functionality to the proposed modified GrabCut approach.